%% file: main.tex
\documentclass{article}

\usepackage{wrapfig}
\usepackage{subcaption} % Add this to your preamble if not already included

\input{packages}
\input{commands}

\title{Understanding Reasoning in Thinking Language Models via Steering Vectors}

\author{Constantin Venhoff$^*$\\
University of Oxford \\
United Kingdom \\
\texttt{constantin@robots.ox.ac.uk} \\
\And Iv\'{a}n Arcuschin$^*$\\
University of Buenos Aires \\
Argentina \\
\texttt{iarcuschin@dc.uba.ar} \\
\AND Philip Torr\\
University of Oxford \\
United Kingdom \\
\And
Arthur Conmy \\
\And Neel Nanda\hspace{0.8cm}\hfill}

\iclrfinalcopy

\begin{document}

\maketitle

\def\thefootnote{*}\footnotetext{Equal contribution}\def\thefootnote{\arabic{footnote}}

\input{abstract}

\input{intro}

\input{preliminaries}

\input{characterizing}

\input{evaluation}

\input{related_work}

\input{conclusion}

\section*{Acknowledgements}

We would like to thank the ML Alignment \& Theory Scholars (MATS) program for supporting this research, and in particular John Teichman and Cameron Holmes for being great research managers. We would also like to thank reviewers from the Workshop on Reasoning and Planning for Large Language Models at ICLR 2025 for extremely helpful feedback on early drafts of this paper.

\bibliography{references}
\bibliographystyle{iclr2025_conference}

\input{appendix}

\end{document}

%% file: packages.tex
\usepackage{iclr2025_conference,times}

\usepackage{microtype}
\usepackage{graphbox} %loads graphicx package
\usepackage{paralist}
\usepackage{booktabs} % for professional tables
\usepackage{colortbl} % For coloring rows in tables
\usepackage{makecell} % For customizing table cells
\usepackage[normalem]{ulem} % for strikethrough

\usepackage{hyperref}

\usepackage{amsmath}
\usepackage{amssymb}
\usepackage{mathtools}
\usepackage{amsthm}

\usepackage[capitalize,noabbrev]{cleveref}

\theoremstyle{plain}

\theoremstyle{definition}

\theoremstyle{remark}

\usepackage[disable,textsize=tiny]{todonotes}
\usepackage{boxedminipage}

\usepackage{soul} % for the command \hl

\usepackage{listings}
\usepackage{xcolor}

%% file: commands.tex
\newboolean{showcomments}
\setboolean{showcomments}{true} % comment this line to deactivate comments

\ifthenelse{\boolean{showcomments}}
{
	\newcommand{\del}[1]{\textcolor{red}{\sout{#1}}}    % please delete
}{
	\newcommand{\del}[1]{}                              % please delete
	
}

\ifthenelse{\boolean{showcomments}}{
	\newcommand{\nbc}[3]{
		{\colorbox{#3}{\bfseries\sffamily\tiny\textcolor{white}{#1}}}
		{\textcolor{#3}{\sf\footnotesize$\langle$\textit{#2}$\rangle$}}}
}{
	\newcommand{\nbc}[3]{}
	
}

{\smallskip
\noindent
\let\emph=\textbf
\begin{boxedminipage}{\columnwidth}\begin{center}\em}%
{\end{center}\end{boxedminipage}%
\medskip
}

\definecolor{verylightgray}{gray}{0.95}

\definecolor{pastelblue}{RGB}{179,205,224}    % Soft blue for deduction
\definecolor{pastelgreen}{RGB}{144,238,144}   % Light green for adding-knowledge
\definecolor{pastelyellow}{RGB}{255,255,153}  % Soft yellow for example-testing
\definecolor{pastelpurple}{RGB}{230,190,255}  % Light purple for backtracking
\definecolor{pastelorange}{RGB}{255,218,185}  % Peach for uncertainty-estimation
\definecolor{pastelcyan}{RGB}{178,236,242}    % Soft cyan for initializing (replacing the original cyan)

%% file: abstract.tex
\begin{abstract}
    Recent advances in large language models (LLMs) have led to the development of \emph{thinking} language models that generate extensive internal reasoning chains before producing responses. While these models achieve improved performance, controlling their reasoning processes remains challenging. This work presents a steering approach for thinking LLMs by analyzing and manipulating specific reasoning behaviors in DeepSeek-R1-Distill models.
    Through a systematic experiment on 500 tasks across 10 diverse categories, we identify several reasoning behaviors exhibited by thinking models, including expressing uncertainty, generating examples for hypothesis validation, and backtracking in reasoning chains. We demonstrate that these behaviors are mediated by linear directions in the model's activation space and can be controlled using steering vectors. 
    By extracting and applying these vectors, we provide a method to modulate specific aspects of the model's reasoning process, such as its tendency to backtrack or express uncertainty. Our approach offers practical tools for steering reasoning processes in thinking models in a controlled and interpretable manner. We validate our steering method using three DeepSeek-R1-Distill models, demonstrating consistent control across different model architectures.
\end{abstract}

%% file: intro.tex
% A recent trend in the development of large language models (LLMs) has been the outstanding success of \emph{thinking} models, which generate extensive internal Chain-of-Thought (CoT) reasoning \citep{reynolds2021promptprogramminglargelanguage,nye2021workscratchpadsintermediatecomputation,Wei2022ChainOT} before outputting user responses. Examples of these models include OpenAI's o1 \citep{learningToReason} and DeepSeek's R1 \citep{deepseekR1}.
% Although these advances have led to significant breakthroughs in performance \citep{o3ARCPrize}, it is still unclear if there is a fundamental difference in the reasoning processes between thinking and non-thinking models.

\section{Introduction}\label{sec:intro}
A recent trend in the development of large language models (LLMs) has been \emph{thinking} LLMs, which generate extensive internal reasoning chains before producing responses \citep{reynolds2021promptprogramminglargelanguage,nye2021workscratchpadsintermediatecomputation,Wei2022ChainOT}. Examples include OpenAI's o1 \citep{learningToReason} and DeepSeek's R1 \citep{deepseekR1}. These models have achieved remarkable improvements in performance \citep{o3ARCPrize}, yet controlling and understanding their internal reasoning processes remains challenging.

To address this challenge, we develop a steering approach for thinking LLMs by analyzing reasoning behaviors in DeepSeek-R1-Distill models. We focus on several reasoning behaviors exhibited by these models, including their tendency to express uncertainty, backtrack in reasoning chains, and generate examples for hypothesis testing. While these behaviors may not constitute a complete taxonomy of reasoning mechanisms, they provide a practical foundation for developing steering methods.

We investigate whether these reasoning behaviors of thinking LLMs can be directly controlled using \emph{steering vectors}, a method that has been shown to allow precise behavioral control in LLMs \citep{Subramani2022May, Turner2023Aug, Zou2023Oct, Panickssery2023Dec, templeton2024scaling, Arditi2024Jun}. By extracting and applying steering vectors, we provide a means to modulate the internal reasoning dynamics of thinking LLMs in a controlled manner.

In summary, our work presents the following key contributions:
\begin{compactenum}
    \item We develop a method to extract steering vectors for specific reasoning behaviors in DeepSeek-R1-Distill models, enabling precise control over aspects of their thinking process, such as \emph{backtracking} and \emph{uncertainty estimation}.
    \item We demonstrate the effectiveness of our steering approach through an empirical evaluation on over $500$ tasks across $10$ diverse categories, showing consistent control across multiple DeepSeek-R1-Distill model architectures and sizes.
\end{compactenum}

\begin{figure}[!t]
    \includegraphics[width=\linewidth]{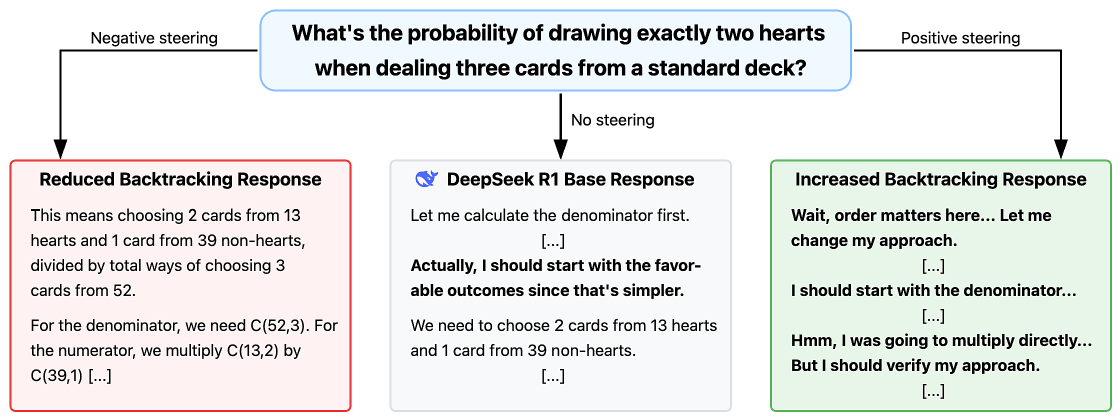}
    \caption{\emph{Steering} on DeepSeek-R1's \textit{backtracking} feature vector changes the model's behavior. Depending on whether we add or subtract this vector to the activations at inference time, the model increases or decreases its tendency to abandon its current approach and explore alternative strategies for the task at hand. Highlighted sections indicate instances of this behavior.}
    \label{fig:steering-example}
\end{figure}
Our steering approach provides a practical method for controlling reasoning processes in thinking models in an interpretable manner, opening new possibilities for fine-grained control of model behavior. To facilitate reproducibility and enable further research, we make our complete experimental codebase and datasets publicly available.\footnote{\url{https://github.com/cvenhoff/steering-thinking-llms}}

%% file: preliminaries.tex
\section{Preliminaries}\label{sec:preliminaries}

\subsection{Attribution Patching}\label{subsec:circuit-discovery}

A fundamental challenge in analyzing the behavior of large language models (LLMs) is identifying the specific components and layers responsible for a given behavior. A widely used technique for addressing this challenge is \textit{activation patching} \citep{meng2022edit}. Activation patching works by replacing the activations of a specific model component with those from a \textit{counterfactual} example, which differs only in a specific aspect of the behavior being analyzed. If this intervention significantly alters the model's output with respect to the observed behavior, the modified component can be marked as playing a key role in implementing this behavior. The \textit{patching effect} is quantified as the change in a relevant output metric:
\[
\Delta L = L(\mathbf{x}_\text{clean} \mid \text{do}(\mathbf{a} = \mathbf{a}_\text{patch})) - L(\mathbf{x}_\text{clean}),
\]
where \( L \) is a metric measuring the difference in model outputs (e.g., KL-divergence), \( \mathbf{a} \) is the original activation, and \( \mathbf{a}_\text{patch} \) is the counterfactual activation.

Since activation patching is computationally expensive, a more efficient linear approximation known as \textit{attribution patching} \citep{nanda2023attribution, syed2023attribution} is often used, which utilizes the gradients of the model's activations with respect to the metric:
\[
\Delta L \approx (\mathbf{a}_\text{patch} - \mathbf{a}_\text{clean})^T \cdot \frac{\partial}{\partial \mathbf{a}_\text{clean}} L(\mathbf{x}_\text{clean} \mid \text{do}(\mathbf{a} = \mathbf{a}_\text{clean})).
\]

\subsection{Computing Steering Vectors}\label{subsec:steering-vectors}

The \textit{Difference of Means} method is a widely used technique for extracting steering vectors in LLMs \citep{turner2024activationadditionsteeringlanguage,arditi2024refusallanguagemodelsmediated}. This technique is based on constructing contrastive datasets that differ in a specific concept and computing the difference in their mean activations of a model.
Formally, let \( D_+ \) and \( D_- \) be two datasets where samples in \( D_+ \) exhibit a given concept, while samples in \( D_- \) do not. Given a model component, we compute the \textit{Difference of Means} vector as:
\[
\mathbf{u} = \frac{1}{|D_+|} \sum_{p_i \in D_+} \mathbf{a}(p_i) - \frac{1}{|D_-|} \sum_{p_j \in D_-} \mathbf{a}(p_j)
\]
where \( a(p_i) \) and \( a(p_j) \) represent the activations of the model components over the prompts from the respective datasets.
This vector \( u \) captures the primary direction in activation space that differentiates the two datasets with respect to the target concept. 
In cases where explicitly matched counterfactuals are unavailable, a common heuristic is to define \( D_+ \) as the set of all samples exhibiting the target behavior, while \( D_- \) consists of the full dataset. In this scenario, the \textit{Difference of Means} vector is computed by subtracting the overall mean activation from the mean activation of the behavior-associated examples.
This isolates the direction in the activation space most associated with the target behavior while reducing the influence of general model biases.

\subsection{DeepSeek Thinking Models}

As mentioned in the introduction, \emph{thinking} models are a type of language model designed to generate long chains of internal reasoning before arriving at a final answer.
Examples of this type of models include QwQ \citep{qwen2024qwq}, Gemini 2.0 Flash Thinking \citep{gemini2024flashthinking}, o1 \citep{learningToReason}, and DeepSeek-R1 \citep{deepseekR1}.

In this work, we focus on characterizing the thinking mechanisms of DeepSeek-R1, a recent \emph{thinking} model that has achieved a similar performance to o1-preview on the ARC-AGI-Pub dataset \citep{r1ArcPrizeAnalysis,o3ARCPrize}.
DeepSeek-R1 is a language model trained through a multi-stage process that combines large-scale reinforcement learning (RL) with the strategic use of supervised fine-tuning (SFT).
The model's architecture uses a Mixture-of-Experts (MoE) approach with 37B activated parameters and 671B total parameters.

The DeepSeek team has distilled R1's reasoning capabilities into smaller, dense models ranging from 1.5B to 70B parameters, based on both Qwen and Llama architectures.
These distilled models achieve similar or better performance than frontier production models like GPT-4o and Claude 3.5 Sonnet at several math and coding benchmarks \citep{deepseekR1}.
We use the \texttt{Qwen-14B} and \texttt{Llama-8B} distilled models of DeepSeek R1 for our analysis.

%% file: characterizing.tex
\section{Identifying Reasoning Behaviors for Steering}\label{sec:characterizing}

\begin{figure}[t]
    \centering
    \includegraphics[width=\linewidth]{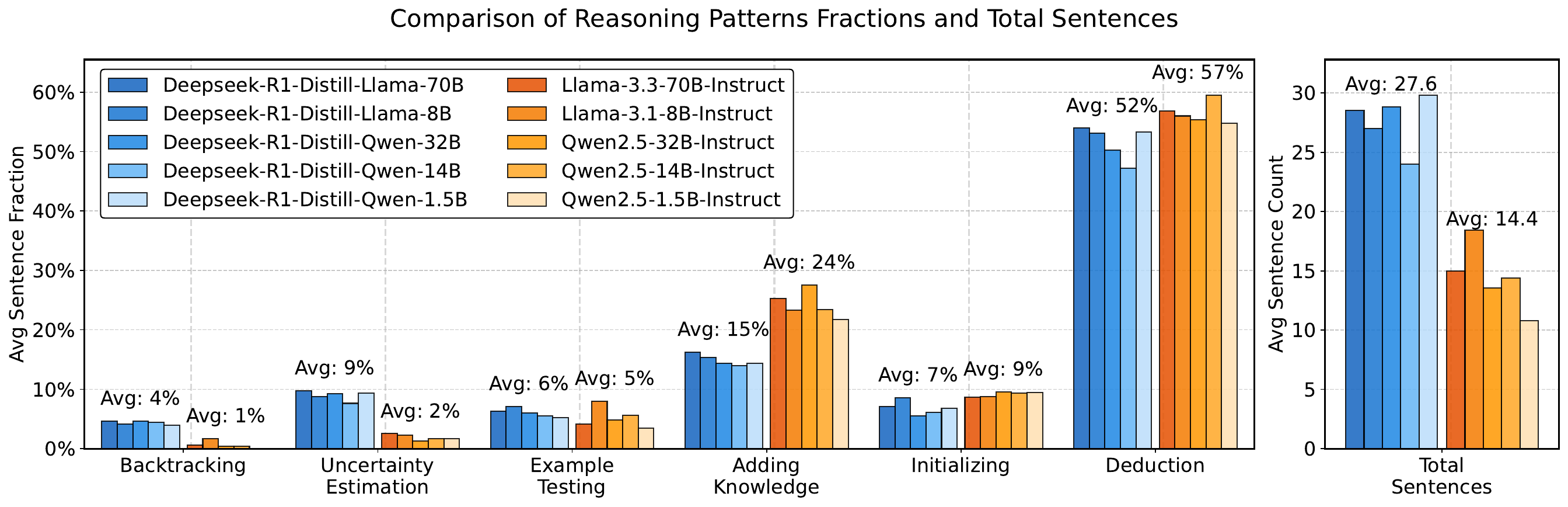}
    \caption{Comparison of behavioral patterns across five DeepSeek-R1-Distill models and five baseline models on $100$ randomly selected tasks from our dataset (cf. \Cref{subsec:setup}). The plot on the left shows the fraction of sentences annotated with each behavioral category. The plot on the right shows the average number of sentences per response. Thinking models generate substantially longer responses ($27.6$ vs $14.4$ sentences on average) and exhibit a higher fractions of backtracking, uncertainty estimation and example testing behaviors,  but lower fractions of knowledge augmentation.}
    \label{fig:baseline-vs-thinking}
\end{figure}

To identify reasoning behaviors suitable for steering, we examined $100$ reasoning chains generated by DeepSeek-R1 and $100$ answers produced by GPT-4o, using tasks across diverse categories (see \cref{subsec:setup}).
Consistent with prior findings \citep{deepseekR1}, we observe that one of the key differences is that the DeepSeek-R1 distills tend to explore many different approaches when solving a task, whereas the base models follow a more linear reasoning trajectory. Notably, the thinking models can express uncertainty about its current approach, generate examples or scenarios to test hypotheses, and backtrack to intermediate steps when revising its reasoning.

Based on these observations, we identify the following behavioral patterns in the DeepSeek-R1 distill reasoning processes as targets for our steering approach:

\begin{compactitem}
    \item \textbf{Initialization}: The model rephrases the task and articulates initial thoughts, typically at the beginning of the reasoning chain.
    \item \textbf{Deduction}: The model derives conclusions based on its current approach and assumptions.
    \item \textbf{Knowledge Augmentation}: The model incorporates external knowledge to refine its reasoning.
    \item \textbf{Example Testing}: The model generates examples or scenarios to validate its working hypothesis.
    \item \textbf{Uncertainty Estimation}: The model explicitly states its confidence or uncertainty regarding its reasoning.
    \item \textbf{Backtracking}: The model abandons its current approach and explores an alternative strategy.
\end{compactitem}

An annotated example of these behaviors is provided in \cref{app:annotated-example}. To systematically annotate reasoning chain in our experiments, we employ GPT-4o (see \cref{app:annotation-process} for details). To quantify the prevalence of these reasoning behaviors, we generate reasoning chains using five DeepSeek-R1-Distill models (Qwen-14B, Qwen-32B, Llama-70B, Qwen-1.5B, and Llama-8B) and compare them against five baseline models (Qwen2.5-Math-1.5B, Qwen2.5-32B-Instruct, Qwen2.5-14B-Instruct, Llama-3.3-70B-Instruct, and Llama-3.1-8B-Instruct) across $100$ diverse tasks (see \cref{subsec:setup}). We then automatically annotate each reasoning chain and compute the fraction of sentences annotated with each behavioral category.
Results are presented in \cref{fig:baseline-vs-thinking}.

The results reveal several key distinctions between thinking and baseline models. Most notably, thinking models generate substantially longer responses, averaging $27.6$ sentences vs $14.4$ for baseline models. Both model types devote similar fractions of their responses to initialization and deduction, suggesting that these processes are comparable across model types. In contrast, baseline models show lower fractions of backtracking, uncertainty estimation, and, to a lesser extent, example testing, instead allocating more of their responses to knowledge augmentation. 

These patterns suggest that uncertainty estimation and backtracking are prominent behaviors that distinguish thinking models, followed by example testing and knowledge addition. 
Given that initialization and deduction do not seem to be behaviors specific to thinking models, steering on these seems conceptually ill-defined;  initialization occurs once at the start and deduction is fundamental to any LLM, therefore we omit them from the empirical study on steering vectors.

%% file: evaluation.tex
\section{Extracting and Evaluating Steering Vectors}\label{sec:evaluation}

In this section, we demonstrate that the reasoning behaviors of the DeepSeek-R1-Distill models characterized in \cref{sec:characterizing} are mediated by linear directions, the \textit{steering vectors}. We assess the causal effect of these vectors by comparing the model's original reasoning chains to those generated under positive and negative steering (adding and subtracting the steering vectors). Our findings indicate that the DeepSeek-R1-Distill models have distinct mechanisms to achieve their reasoning process. Additionally, our steering vectors provide an efficient way to influence these models' reasoning behavior, for example, increasing their tendency to backtrack or modulate their inherent uncertainty in their own reasoning.

\subsection{Experimental Setup}\label{subsec:setup}

For our experiments, we generate a dataset of $500$ tasks across 10 categories using Claude 3.5 Sonnet (see \Cref{tab:dataset-categories}).
We conduct our experiments on three DeepSeek-R1-Distill models: \texttt{Qwen-14B}, \texttt{Qwen-1.5B}, and \texttt{Llama-8B}. When generating a reasoning chain, we use greedy decoding and $1000$ max tokens per response.

\begin{table}[t]
    \rowcolors{2}{white}{gray!25}
    \centering
    \begin{tabular}{p{0.25\linewidth}p{0.65\linewidth}}
        \toprule
        \textbf{Category} & \textbf{Description} \\
        \midrule
        Mathematical Logic & Problems requiring formal logical operations, mathematical proofs, and numerical reasoning \\
        Spatial Reasoning & Tasks involving visualization, geometric manipulation, and understanding spatial relationships \\
        Verbal Logic & Problems focused on language-based reasoning, syllogisms, and verbal analogies \\
        Pattern Recognition & Questions requiring identification and continuation of sequences or abstract patterns \\
        Lateral Thinking & Problems that require creative, non-linear approaches to reach unconventional solutions \\
        Causal Reasoning & Tasks involving understanding cause-and-effect relationships and making causal inferences \\
        Probabilistic Thinking & Problems requiring reasoning about uncertainty, probability, and statistical concepts \\
        Systems Thinking & Questions about complex systems, interconnected components, and emergent behaviors \\
        Creative Problem Solving & Open-ended problems requiring novel approaches and innovative solutions \\
        Scientific Reasoning & Tasks involving hypothesis formation, experimental design, and evidence evaluation \\
        \bottomrule
    \end{tabular}
    \caption{Task categories used to analyze reasoning behaviors.}
    \label{tab:dataset-categories}
\end{table}

\subsection{Locating Causally Relevant Activations}\label{subsec:locating-relevant-activations}

To extract robust steering vectors, we first identify the activations where these vectors are linearly represented within the model. We focus on the \textit{residual stream activations}, i.e., the outputs of each transformer layer. Given a reasoning chain generated by a DeepSeek-R1-Distill model, we identify both the \textit{token positions} and \textit{layers} where the steering vector is active. This process consists of two key steps:

\begin{enumerate}
    \item \textbf{Identifying relevant token positions}: Determine which tokens in the reasoning chain correspond to a specific behavioral category.
    \item \textbf{Determining causally relevant layers}: Use \textit{attribution patching} (\cref{sec:preliminaries}) to evaluate which layers contribute causally to the behavior in question.
\end{enumerate}

To obtain token positions associated with each behavioral category, we generate $500$ reasoning chains with the tasks introduced in \cref{subsec:setup}, using both DeepSeek-R1-Distill models and then annotate them automatically with GPT-4o. Since the DeepSeek-R1-Distill models are autoregressive, we consider for each category both the token position \textbf{preceding} the start of a token-sequence annotated with the current category and the annotated sequence itself as the causally relevant token positions (up to $10$ tokens). This ensures that we capture both the decision point where the model transitions into the behavior and the behavior's execution phase.

To identify the causally relevant layers for each behavioral category, we first extract a steering vector candidate from every layer using the \textit{Difference of Means} method (\cref{subsec:steering-vectors}):
\[
\mathbf{u}_\ell^c = \frac{1}{|D_+|} \sum_{p_i \in D_+} \mathbf{\bar{a}}_\ell^c (p_i) - \frac{1}{|D_-|} \sum_{p_j \in D_-} \mathbf{a}_\ell^c (p_j), 
\quad \text{with} \quad \mathbf{\bar{a}}_\ell^c (p_i) = \frac{1}{|\text{seq}_c(p_i)|} \sum_{t \in \text{seq}_c(p_i)} \mathbf{a}_\ell (t).
\]
where:
\begin{itemize}
    \item \( \mathbf{a}_\ell (t) \) represents the residual stream activation at layer \( \ell \) for token position \( t \).
    \item \( \text{seq}_c(p) \) is the set of all token sequences within prompt \( p \) that are annotated with category \( c \), including the preceding token position.
    \item \( \mathbf{\bar{a}}_\ell^c (p_i) \) denotes the \textit{mean activation} across all token positions within the annotated sequences of category \( c \) at layer \( \ell \).
    \item \( D_+ \) consists of prompts containing at least one sequence labeled with category \( c \), while \( D_- \) represents the full dataset.
\end{itemize}

The resulting vector \( \mathbf{u}_\ell^c \) serves as a candidate steering vector for each layer. To ensure consistent scaling across different sequence lengths and behaviors, we normalize each steering vector to have the same magnitude as the mean overall activation:
\[
\mathbf{u}_\ell^{c,\text{norm}} = \mathbf{u}_\ell^c \cdot \frac{\|\mathbf{\bar{a}}_\ell^\text{overall}\|}{\|\mathbf{u}_\ell^c\|}
\]
where \( \mathbf{\bar{a}}_\ell^\text{overall} \) is the mean activation across all tokens in the dataset at layer \( \ell \). This normalization helps account for varying sequence lengths and ensures steering vectors have comparable magnitudes across different behavioral categories.

\subsection{Selecting Final Steering Vectors}\label{subsec:selecting-steering-vectors}
\begin{figure}[!t]
    \centering
    \begin{subfigure}[b]{\linewidth}
        \centering
        \includegraphics[width=\linewidth]{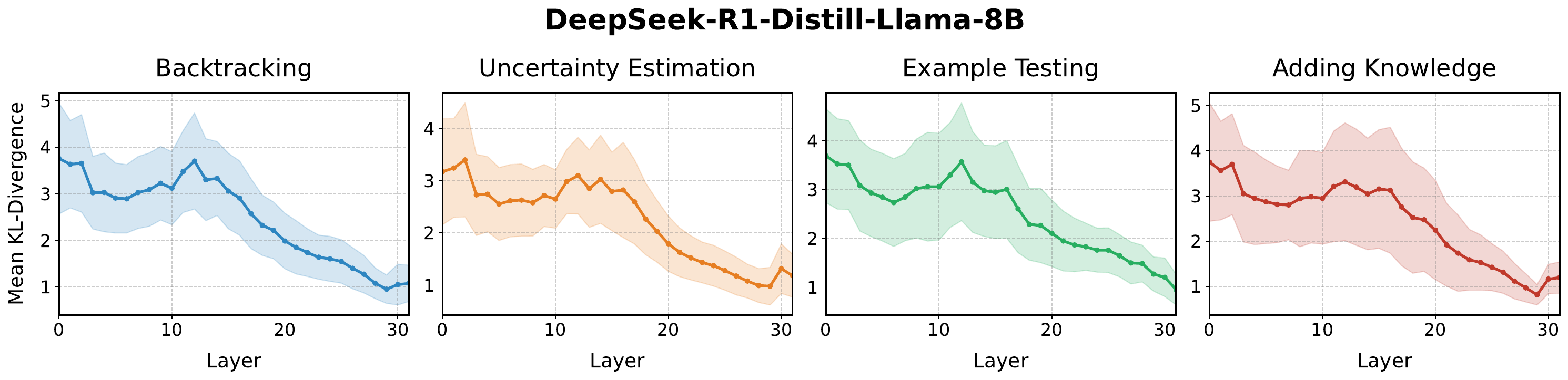}
        \label{fig:attribution-effects-1}
    \end{subfigure}\vspace{-0.2cm}
    \begin{subfigure}[b]{\linewidth}
        \centering
        \includegraphics[width=\linewidth]{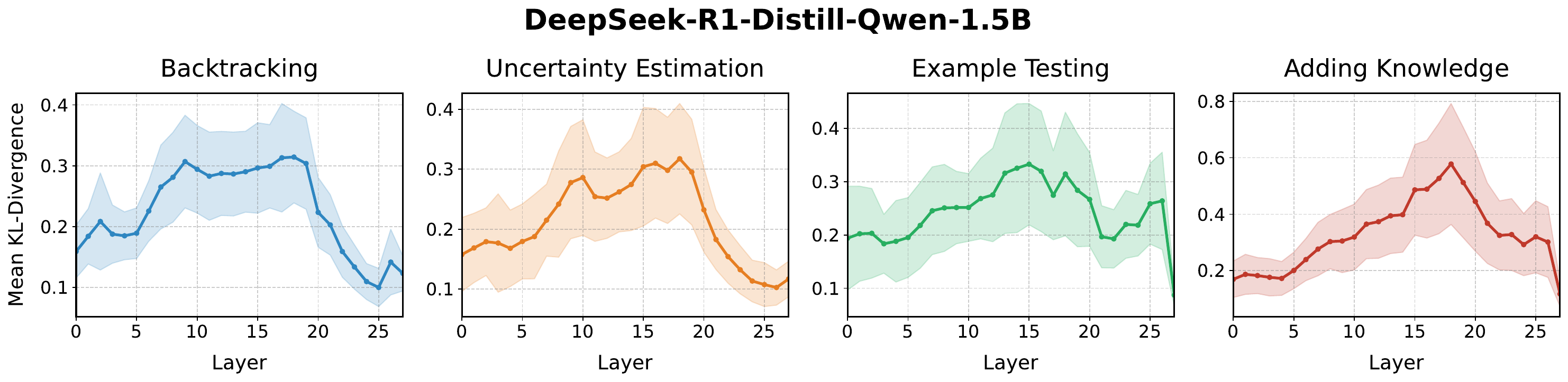}
        \label{fig:attribution-effects-2}
    \end{subfigure}\vspace{-0.2cm}
    \begin{subfigure}[b]{\linewidth}
        \centering
        \includegraphics[width=\linewidth]{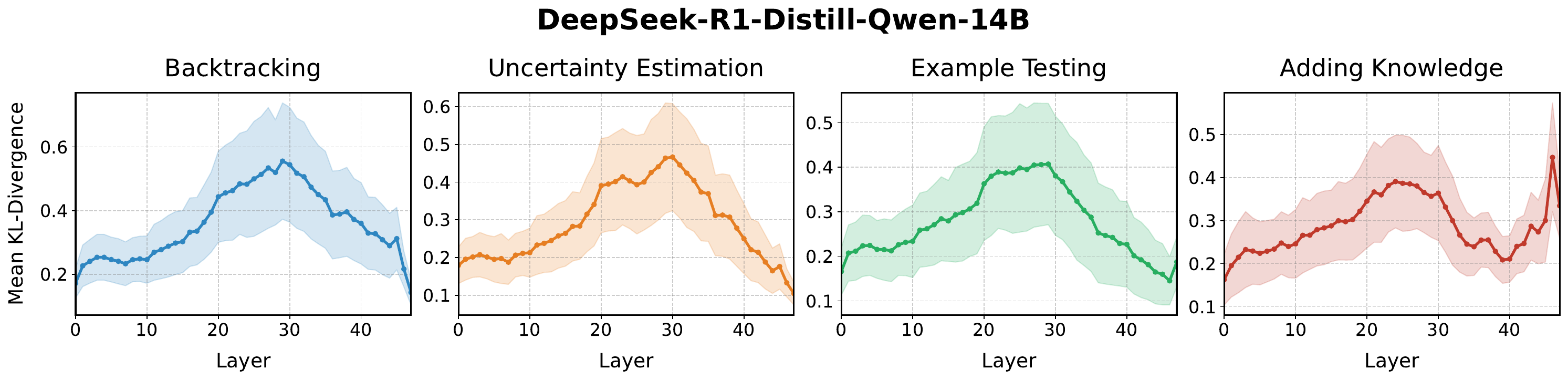}
        \label{fig:attribution-effects-3}
    \end{subfigure}
    \caption{Causal impact of candidate steering vectors across model layers. The y-axis represents the absolute mean KL-divergence for the next-token logit distribution when removing the steering vector at each layer. The steering vectors for all reasoning mechanisms have similar peaks in the middle layers of the respective models.}
    \label{fig:attribution-effects}
\end{figure}

To determine the final steering vectors, we apply \textit{attribution patching} (\cref{subsec:circuit-discovery}) to quantify the causal relevance of each vector in its respective layer. Specifically, we consider the following patching experiment: Given a candidate steering vector \( \mathbf{u}_\ell^c \) for a specific behavioral category, we add it to the residual stream activation preceding a token-sequence annotated with one of the relevant behaviors. Therefore, we define the patched activation as:
\[
\mathbf{a}_\ell^\text{patched} = \mathbf{a}_\ell + \mathbf{u}_\ell^c.
\]
If this intervention leads to a significant change in the KL divergence of the next-token prediction, then the steering vector in layer \( \ell \) is \textit{causally relevant} for the given behavior. We approximate the patching effect for this experiment with:
\[
\Delta L \approx (\mathbf{u}_\ell^c)^T \cdot \frac{\partial}{\partial \mathbf{a}_\ell} L(\mathbf{x}_\text{clean} \mid \text{do}(\mathbf{a}_\ell = \mathbf{a}_\text{clean})),
\]
where \( \mathbf{u}_\ell^c = (\mathbf{a}_\ell^\text{patched} - \mathbf{a}_\ell) \). 
We average the absolute patching effect for each category over all category-sequences in all $500$ reasoning chains. The results are shown in \cref{fig:attribution-effects}. Based on these results, we can select the causally most relevant steering vectors from the layers where the patching scores are highest. However, we avoid selecting early layers that show excessive correlation with embedding tokens, as these layers primarily capture token-specific representations rather than behavioral patterns (\cref{app:embedding-similarity}).

\subsection{Evaluation of Steering Vectors}\label{subsec:evaluating-steering-vectors}

To evaluate the effectiveness of our extracted steering vectors, we apply them at the selected layers (see \cref{tab:selected-layers}) and observe their influence on the model's reasoning process. Steering is implemented by \textit{adding or subtracting} the extracted steering vectors \( \mathbf{u}_\ell^c \) to the residual stream activations at inference time.
By applying this intervention, we can increase or decrease behaviors such as backtracking, uncertainty estimation, and example testing, providing a direct mechanism for manipulating the model's reasoning process.

\subsection{Results of Steering}\label{sec:steering-results}
\begin{figure}[t]
    \centering
    \includegraphics[width=\linewidth]{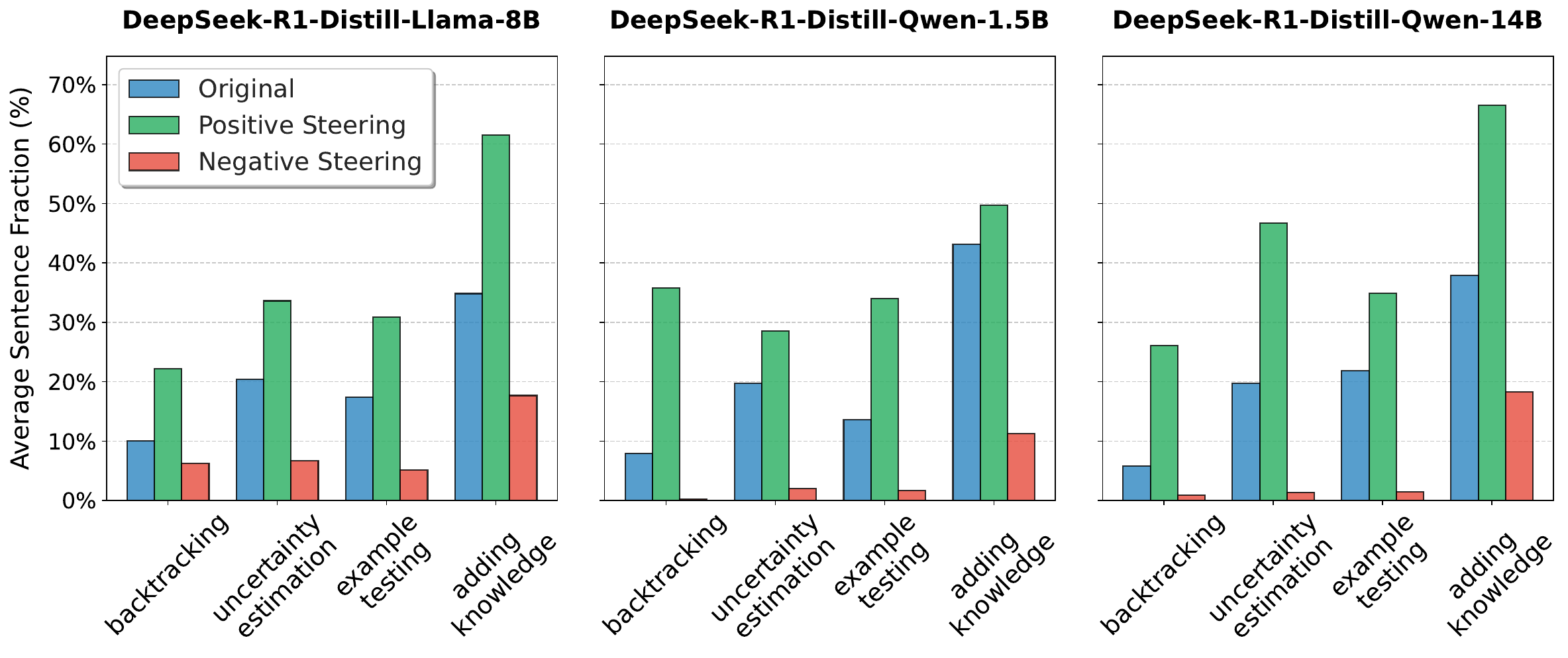}
    \caption{Effect of applying the steering vector for each reasoning behavior across different distill models. The y-axis shows the change in the fraction of tokens exhibiting each behavior when applying positive or negative steering. Positive steering increases behaviors such as backtracking and uncertainty estimation, while negative steering suppresses or significantly reduces them, confirming the causal influence of our extracted vectors.}
    \label{fig:steering-results}
\end{figure}

\begin{table}[t]
    \rowcolors{2}{white}{gray!25}
    \centering
    \begin{tabular}{lccc}
        \toprule
        \textbf{Behavioral Category} & \textbf{\shortstack[c]{DeepSeek-R1-Distill \\ Llama-8B}} & \textbf{\shortstack[c]{DeepSeek-R1-Distill \\ Qwen-1.5B}} & \textbf{\shortstack[c]{DeepSeek-R1-Distill \\ Qwen-14B}} \\
        \midrule
        Uncertainty Estimation & 12 & 18 & 29 \\
        Example Testing & 12 & 15 & 29 \\
        Backtracking & 12 & 17 & 29 \\
        Adding Knowledge & 12 & 18 & 24 \\
        \bottomrule
    \end{tabular}
    \caption{Selected layers for each behavioral category based on attribution patching results. For each model, we select the layer with the maximum score from the attribution patching experiments, ignoring early layers that are highly correlated with embedding tokens.}
    \label{tab:selected-layers}
\end{table}

We apply each steering vector to $50$ unseen reasoning tasks and analyze how the model's reasoning behavior changes. The results, presented in \cref{fig:steering-results}, demonstrate that our extracted vectors effectively control the model's reasoning patterns.
\Cref{app:example-steering} includes a full example of positive and negative steering on the ``Adding Knowledge'' vector for the DeepSeek-R1-Distill-Qwen-14B model.

As shown in \cref{fig:steering-results}, positive steering increases behaviors such as backtracking, uncertainty estimation, and example testing, while negative steering reduces them. These effects are consistent across both DeepSeek-R1-Distill models, reinforcing the hypothesis that Thinking LLMs encode these reasoning mechanisms as linear directions in their activation space. Our findings confirm that steering vectors provide a reliable and efficient method for interpreting and controlling the internal reasoning dynamics of thinking large language models. Further analysis of the cosine similarity between different steering vectors reveals that most behavioral categories correspond to distinct directions in the model's activation space (see \cref{app:cosine-similarity}).

%% file: related_work.tex
\section{Related Work}\label{sec:related}

Recent work has explored methods for steering and interpreting language models by identifying meaningful directions or features within their internal representation spaces. \citet{Subramani2022May} show that extracting latent steering vectors from pretrained language models can systematically alter the generated text. Similarly, \citet{Turner2023Aug} propose \emph{activation engineering}, modifying model activations at inference time to control outputs, in contrast to prompt engineering or fine-tuning. Extending this line of research, \citet{Panickssery2023Dec} introduce Contrastive Activation Addition (CAA), which derives a “contrastive” vector by averaging differences in residual stream activations between positive and negative examples of a target behavior. Adding this vector to a model's activations elicits more desirable outputs without retraining. Beyond these methods, \citet{Zou2023Oct} propose \emph{Representation Engineering}, offering a top-down approach to refining and analyzing internal representations for greater transparency.

Applying these methods for fine-grained control of language models through systematic manipulation of their internal representations \citet{Li2023Jun} propose an \emph{inference-time intervention} to encourage more truthful responses without additional training. Additionally, \citet{Arditi2024Jun} has shown that refusal behavior can be localized to a single direction in latent space, that minimally affect other capabilities, enabling targeted interventions to encourage refusal or jailbreak the model.

A related line of work focuses on leveraging these internal representations for reasoning and chain-of-thought. \citet{Zhao2025ActivationControlEfficientlyEliciting} found a steering vector to efficiently elicit long chain-of-thought in language models. \citet{dutta2024thinkstepbystepmechanisticunderstanding} analyzed the neural sub-structures within Llama-2 7B that facilitate multistep reasoning over fictional ontologies and found that the model deploys multiple parallel pathways for step-by-step reasoning. In concurrent work, \citet{hazra2025under} train sparse autoencoders on DeepSeek's 671B-parameter R1 reasoning model, uncovering internal features associated with reasoning behaviors, such as backtracking, which can be used for steering interventions.

%% file: conclusion.tex
\section{Conclusion and Future Work}

This work presents a steering approach for controlling reasoning behaviors in thinking LLMs, with a specific focus on DeepSeek-R1-Distill models. We demonstrate that several reasoning behaviors exhibited by these models, including expressing uncertainty, backtracking, and example testing, can be effectively controlled using steering vectors. While our analysis does not claim to provide a complete taxonomy of reasoning mechanisms, it establishes a practical framework for steering specific aspects of model behavior. 
Our key findings indicate that:
\begin{itemize}
    \item Several reasoning behaviors in thinking models can be isolated to specific directions in the model's activation space, enabling precise control through steering vectors.
    \item Our steering approach is effective across a diverse set of 500 tasks, demonstrating robust control over targeted reasoning behaviors.
    \item The method generalizes across different model architectures within the DeepSeek-R1-Distill family, showing consistent steering effects.
\end{itemize}
These results provide practical tools for modulating reasoning capabilities in thinking models. The ability to adjust specific aspects of the reasoning process through steering vectors opens new possibilities for adapting these models to different tasks and requirements.

Despite these promising results, our work has several limitations that suggest directions for future research. The current automated annotation process using GPT-4o, while efficient, has occasionally produced false positives and false negatives in identifying reasoning patterns. Future work should focus on developing more robust annotation methods, potentially incorporating multiple models or human validation to improve accuracy. Additionally, while our analysis centers on DeepSeek-R1-Distill models, the generalization of these findings to other models that have undergone reward-based RL training instead of being fine-tuned on thinking models (e.g., Qwen's QwQ) remains an open question. Extending this research to a broader range of models would provide deeper insights into the universality of thinking mechanisms and their practical applications.
By addressing these limitations, future research can further advance the understanding and control of thinking models, paving the way for more reliable and adaptable AI systems.

%% file: appendix.tex
\newpage
\appendix
\onecolumn

\crefalias{section}{appendix}
\crefalias{subsection}{appendix}
\crefalias{subsubsection}{appendix}

\section{Details on the Annotation Process}\label{app:annotation-process}

We use the following prompt to automatically annotate LLM responses:

\begin{lstlisting}[breakindent=0pt]
Please split the following reasoning chain of an LLM into annotated parts using labels and the following format ["label"]...["end-section"]. A sentence should be split into multiple parts if it incorporates multiple behaviours indicated by the labels.

Available labels:
0. initializing -> The model is rephrasing the given task and states initial thoughts.
1. deduction -> The model is performing a deduction step based on its current approach and assumptions.
2. adding-knowledge -> The model is enriching the current approach with recalled facts.
3. example-testing -> The model generates examples to test its current approach.
4. uncertainty-estimation -> The model is stating its own uncertainty.
5. backtracking -> The model decides to change its approach.

The reasoning chain to analyze:
{thinking_process}

Answer only with the annotated text. Only use the labels outlined above. If there is a tail that has no annotation leave it out.
\end{lstlisting}

\section{Annotated Example}\label{app:annotated-example}

\textbf{Riddle}: What has cities, but no houses; forests, but no trees; and rivers, but no water?

\textbf{DeepSeek R1 Response}:
\begin{lstlisting}[breakindent=0pt]
Okay, so I came across this riddle. At first glance, it seems tricky, but I can break it down. 
Let me analyze each part separately. The first part: "What has cities, but no houses." The answer must be something that contains cities but lacks houses. Cities have buildings, roads, and infrastructure, but without houses, it must be something that represents a city rather than an actual one.
Maybe it's a city without residential buildings, but that doesn't fit perfectly.
Wait, perhaps it refers to something that *displays* cities rather than being a real city.
What about the second part: "forests, but no trees"? Forests are made of trees, so if there are no trees, it isn't really a forest. But perhaps it's something that symbolizes a forest without actually being one.
Finally, "rivers, but no water." Rivers are defined by flowing water, so if there's no water, it might be referring to a representation of a river rather than a real one.
Considering all three parts together, the answer must be something that includes symbolic representations of cities, forests, and rivers but lacks their real-world components. A map fits this description-it has cities, forests, and rivers marked on it, but they are not real.
\end{lstlisting}

\textbf{Automatically annotated response}: (colored by assigned label)
\begin{lstlisting}[breakindent=0pt]
(*@\sethlcolor{pastelcyan}\hl{["initializing"]}@*)(*@\sethlcolor{pastelcyan}\hl{Okay, so I came across this riddle. At first glance, it seems tricky, but I can break it down.}@*)(*@\sethlcolor{pastelcyan}\hl{["end-section"]}@*)
(*@\sethlcolor{pastelblue}\hl{["deduction"]}@*)(*@\sethlcolor{pastelblue}\hl{Let me analyze each part separately. The first part: "What has cities, but no houses." The answer must be something that contains cities but lacks houses.}@*)(*@\sethlcolor{pastelblue}\hl{["end-section"]}@*)
(*@\sethlcolor{pastelgreen}\hl{["adding-knowledge"]}@*)(*@\sethlcolor{pastelgreen}\hl{Cities have buildings, roads, and infrastructure, but without houses, it must be something that represents a city rather than an actual one.}@*)(*@\sethlcolor{pastelgreen}\hl{["end-section"]}@*)
(*@\sethlcolor{pastelyellow}\hl{["example-testing"]}@*)(*@\sethlcolor{pastelyellow}\hl{Maybe it's a city without residential buildings, but that doesn't fit perfectly.}@*)(*@\sethlcolor{pastelyellow}\hl{["end-section"]}@*)
(*@\sethlcolor{pastelpurple}\hl{["backtracking"]}@*)(*@\sethlcolor{pastelpurple}\hl{Wait, perhaps it refers to something that *displays* cities rather than being a real city.}@*)(*@\sethlcolor{pastelpurple}\hl{["end-section"]}@*)
(*@\sethlcolor{pastelblue}\hl{["deduction"]}@*)(*@\sethlcolor{pastelblue}\hl{What about the second part: "forests, but no trees"?}@*)(*@\sethlcolor{pastelblue}\hl{["end-section"]}@*)
(*@\sethlcolor{pastelgreen}\hl{["adding-knowledge"]}@*)(*@\sethlcolor{pastelgreen}\hl{Forests are made of trees, so if there are no trees, it isn't really a forest.}@*)(*@\sethlcolor{pastelgreen}\hl{["end-section"]}@*)
(*@\sethlcolor{pastelyellow}\hl{["example-testing"]}@*)(*@\sethlcolor{pastelyellow}\hl{Unless it's an area that used to be a forest but was deforested-but that wouldn't be called a forest anymore.}@*)(*@\sethlcolor{pastelyellow}\hl{["end-section"]}@*)
(*@\sethlcolor{pastelorange}\hl{["uncertainty-estimation"]}@*)(*@\sethlcolor{pastelorange}\hl{That might be a stretch, though.}@*)(*@\sethlcolor{pastelorange}\hl{["end-section"]}@*)
(*@\sethlcolor{pastelblue}\hl{["deduction"]}@*)(*@\sethlcolor{pastelblue}\hl{The third part: "rivers, but no water."}@*)(*@\sethlcolor{pastelblue}\hl{["end-section"]}@*)
(*@\sethlcolor{pastelgreen}\hl{["adding-knowledge"]}@*)(*@\sethlcolor{pastelgreen}\hl{Rivers are defined by flowing water. If there's no water, it might be referring to a representation of a river rather than a real one.}@*)(*@\sethlcolor{pastelgreen}\hl{["end-section"]}@*)
(*@\sethlcolor{pastelyellow}\hl{["example-testing"]}@*)(*@\sethlcolor{pastelyellow}\hl{Maybe it's a dry riverbed, but then it's not technically a river anymore.}@*)(*@\sethlcolor{pastelyellow}\hl{["end-section"]}@*)
(*@\sethlcolor{pastelblue}\hl{["deduction"]}@*)(*@\sethlcolor{pastelblue}\hl{Putting it all together, the answer must be something that includes symbolic representations of cities, forests, and rivers but lacks their real-world components.}@*)(*@\sethlcolor{pastelblue}\hl{["end-section"]}@*)
(*@\sethlcolor{pastelblue}\hl{["deduction"]}@*)(*@\sethlcolor{pastelblue}\hl{A map fits this description-it has cities, forests, and rivers marked on it, but they are not real.}@*)(*@\sethlcolor{pastelblue}\hl{["end-section"]}@*)
\end{lstlisting}

\section{Cosine Similarity Between Feature Vectors}\label{app:cosine-similarity}

To better understand the relationships between different reasoning behaviors in thinking models, we analyze the cosine similarity between the extracted steering vectors for different behavioral categories. This analysis provides insights into how distinct these reasoning mechanisms are in the model's representational space and whether certain behaviors share similar underlying features.

We compute pairwise cosine similarities between the steering vectors for five key behavioral categories: \textit{initializing}, \textit{backtracking}, \textit{uncertainty-estimation}, \textit{adding-knowledge}, and \textit{deduction}. For each pair of behavioral categories, we calculate the cosine similarity between their corresponding steering vectors at the layers identified as causally relevant through attribution patching (\cref{subsec:selecting-steering-vectors}).

The cosine similarity between two feature vectors $\mathbf{u}^{c_1}$ and $\mathbf{u}^{c_2}$ is computed as:
\[
\text{sim}(\mathbf{u}^{c_1}, \mathbf{u}^{c_2}) = \frac{\mathbf{u}^{c_1} \cdot \mathbf{u}^{c_2}}{|\mathbf{u}^{c_1}| |\mathbf{u}^{c_2}|}
\]

\begin{figure}[!t]
    \centering
    \begin{subfigure}[b]{0.49\linewidth}
        \centering
        \includegraphics[width=\linewidth]{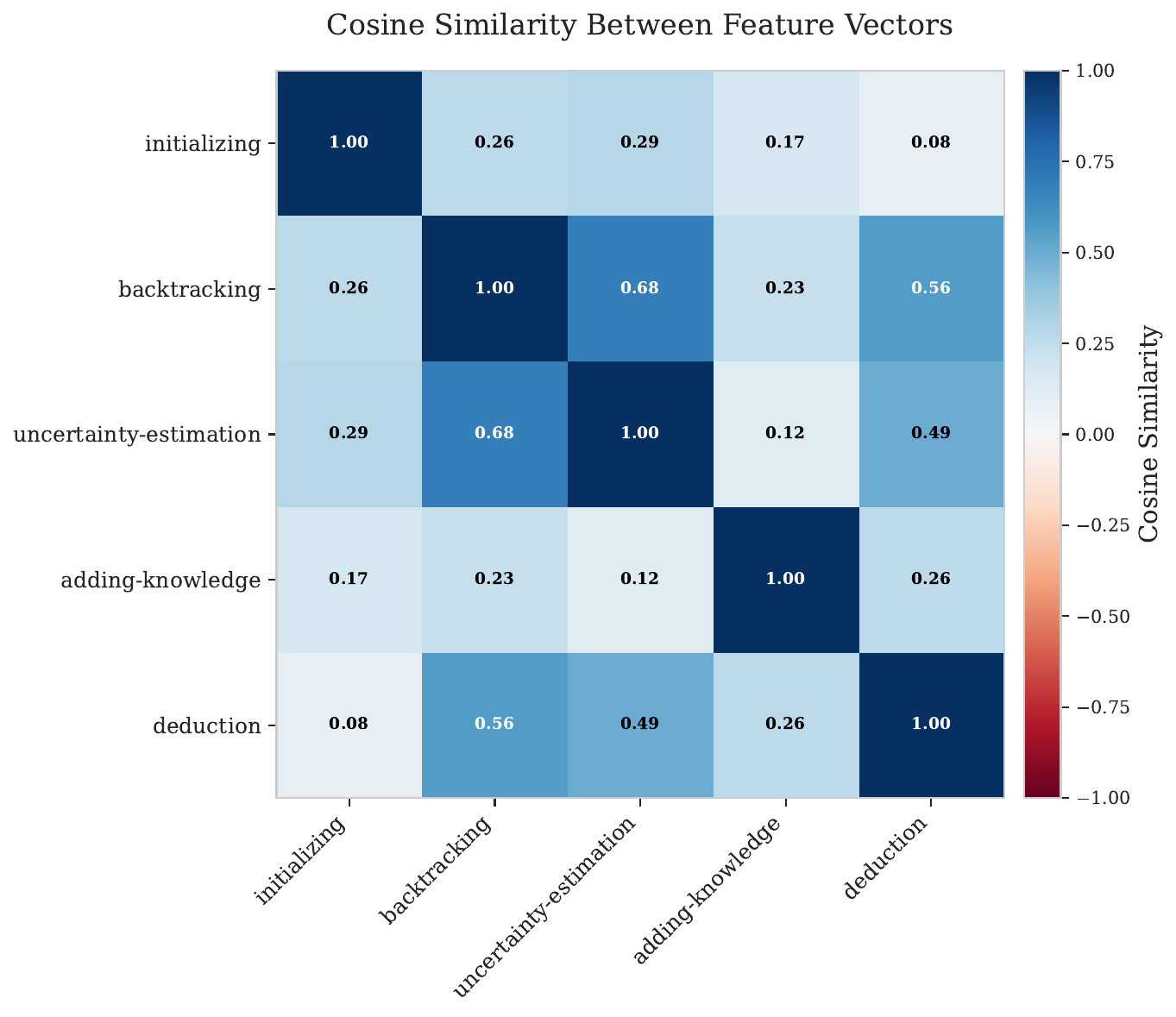}
        \caption{DeepSeek-R1-Distill-Llama-8B}
        \label{fig:cosine-similarity-llama}
    \end{subfigure}
    \hfill
    \begin{subfigure}[b]{0.49\linewidth}
        \centering
        \includegraphics[width=\linewidth]{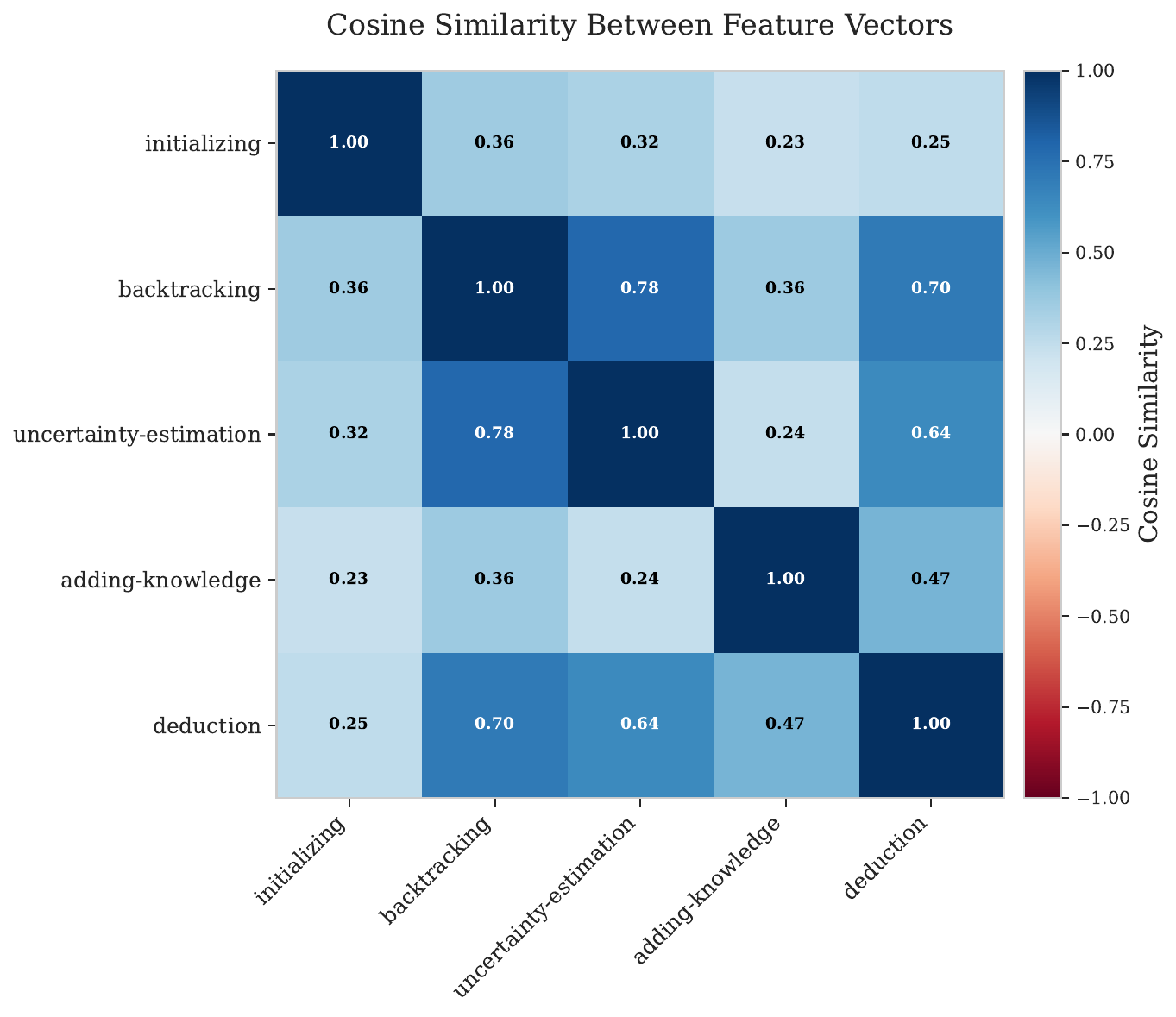}
        \caption{DeepSeek-R1-Distill-Qwen-14B}
        \label{fig:cosine-similarity-qwen}
    \end{subfigure}
    \caption{Cosine similarity heatmaps between steering vectors for different reasoning behaviors. The heatmaps show pairwise similarities between feature vectors extracted for five behavioral categories. Values range from -1 (completely opposite) to 1 (identical direction), with colors indicating the strength and direction of similarity. Most behaviors show low to moderate similarities, indicating they represent distinct reasoning mechanisms in the model's activation space.}
    \label{fig:cosine-similarity-heatmaps}
\end{figure}

\cref{fig:cosine-similarity-heatmaps} presents the cosine similarity heatmaps for both DeepSeek-R1-Distill models. Several key observations emerge from this analysis:

\begin{enumerate}
    \item \textit{Distinct reasoning mechanisms}: Most behavioral categories show relatively low cosine similarities with each other, indicating that they correspond to distinct directions in the model's activation space. This supports our hypothesis that different reasoning behaviors are mediated by separate linear directions.
    
    \item \textit{Uncertainty and backtracking correlation}: Both models show moderate positive correlation between \textit{uncertainty-estimation} and \textit{backtracking} behaviors, which aligns with our intuition that models often express uncertainty before deciding to change their reasoning approach. Despite this cosine similarity overlap, our steering experiments (see \cref{sec:steering-results}) demonstrate that these represent fundamentally different mechanisms in the model's reasoning process.
    
    \item \textit{Model-specific patterns}: While the overall structure is similar across models, there are notable differences in the specific similarity values, suggesting that different model architectures may encode these reasoning behaviors with varying degrees of separation.
\end{enumerate}

These results demonstrate that the extracted steering vectors capture meaningful and largely orthogonal directions in the model's representational space, validating our approach for isolating specific reasoning mechanisms in thinking language models.

\section{Cosine Similarity with Embedding and Unembedding Vectors}\label{app:embedding-similarity}

To better understand the relationship between our extracted steering vectors and the model's embedding space, we analyze the cosine similarity between steering vectors and the model's embedding and unembedding matrices across different layers. This analysis helps explain why certain layers are more effective for steering than others.

\subsection{Methodology}

We compute the cosine similarity between each steering vector at each layer and the model's embedding matrix (\texttt{model.embed\_tokens.weight}) and unembedding matrix (\texttt{model.lm\_head.weight}). For each steering vector $\mathbf{u}_\ell^c$ at layer $\ell$ and behavioral category $c$, we calculate:

\begin{align}
\text{sim}_{\text{embed}}(\mathbf{u}_\ell^c) &= \max_i \cos(\mathbf{u}_\ell^c, \mathbf{E}_i) \\
\text{sim}_{\text{unembed}}(\mathbf{u}_\ell^c) &= \max_j \cos(\mathbf{u}_\ell^c, \mathbf{U}_j)
\end{align}

where $\mathbf{E}_i$ represents the $i$-th row of the embedding matrix and $\mathbf{U}_j$ represents the $j$-th row of the unembedding matrix.

\subsection{Results and Analysis}

\begin{figure}[!t]
    \centering
    \begin{subfigure}[b]{0.49\linewidth}
        \centering
        \includegraphics[width=\linewidth]{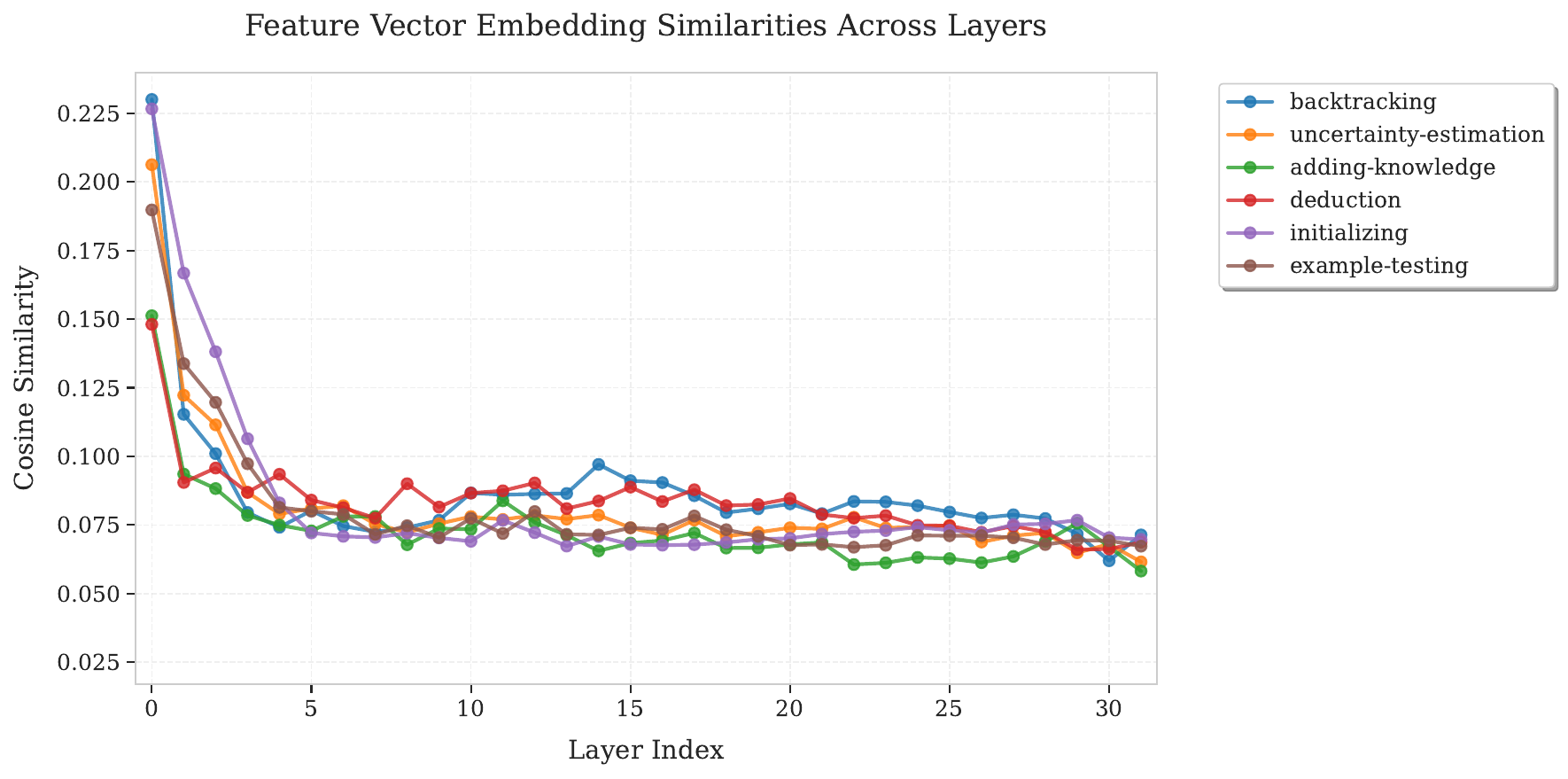}
        \caption{Embedding similarities - Llama-8B}
        \label{fig:embedding-similarity-llama}
    \end{subfigure}
    \hfill
    \begin{subfigure}[b]{0.49\linewidth}
        \centering
        \includegraphics[width=\linewidth]{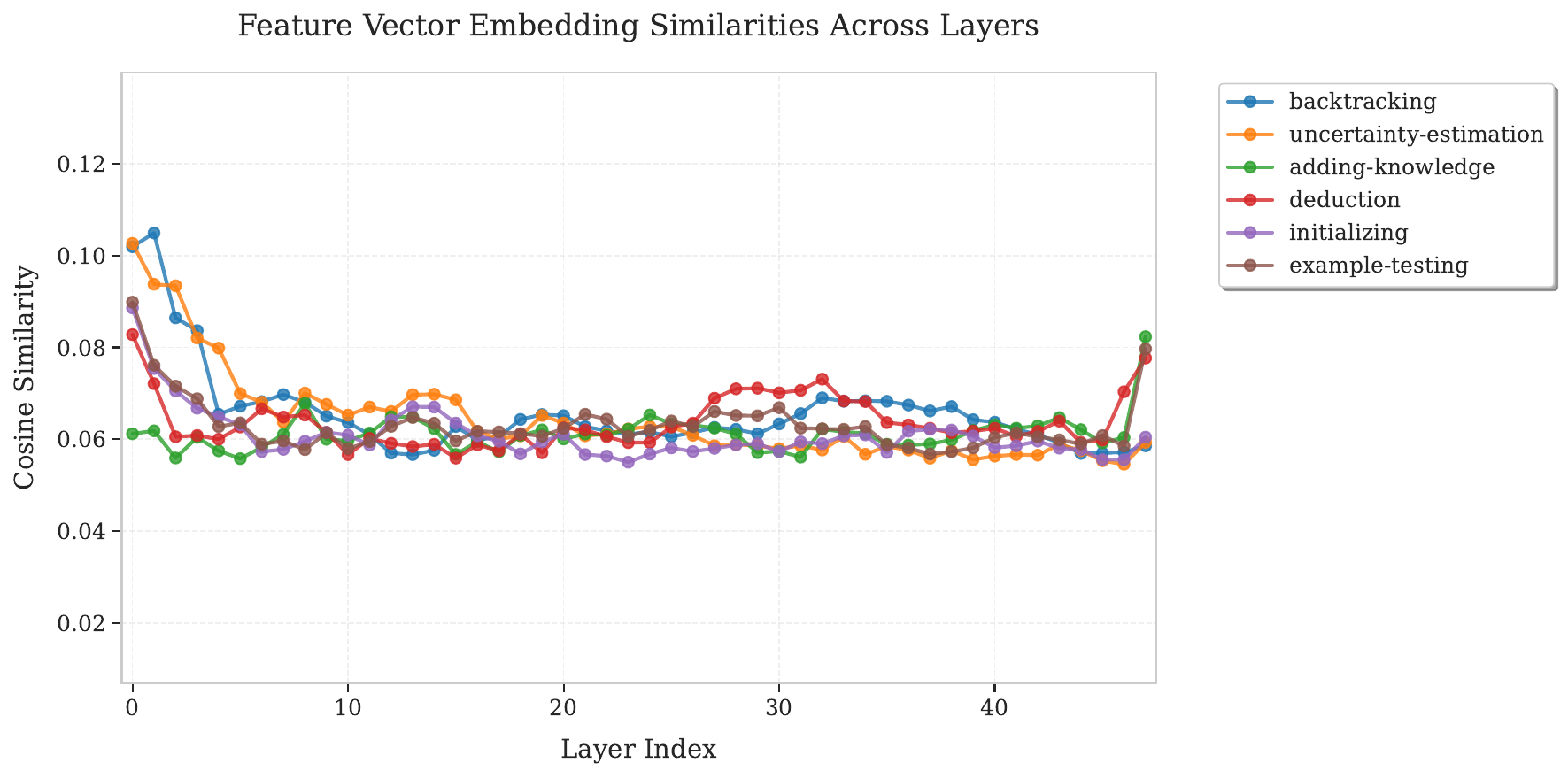}
        \caption{Embedding similarities - Qwen-14B}
        \label{fig:embedding-similarity-qwen}
    \end{subfigure}
    
    \vspace{0.5cm}
    
    \begin{subfigure}[b]{0.49\linewidth}
        \centering
        \includegraphics[width=\linewidth]{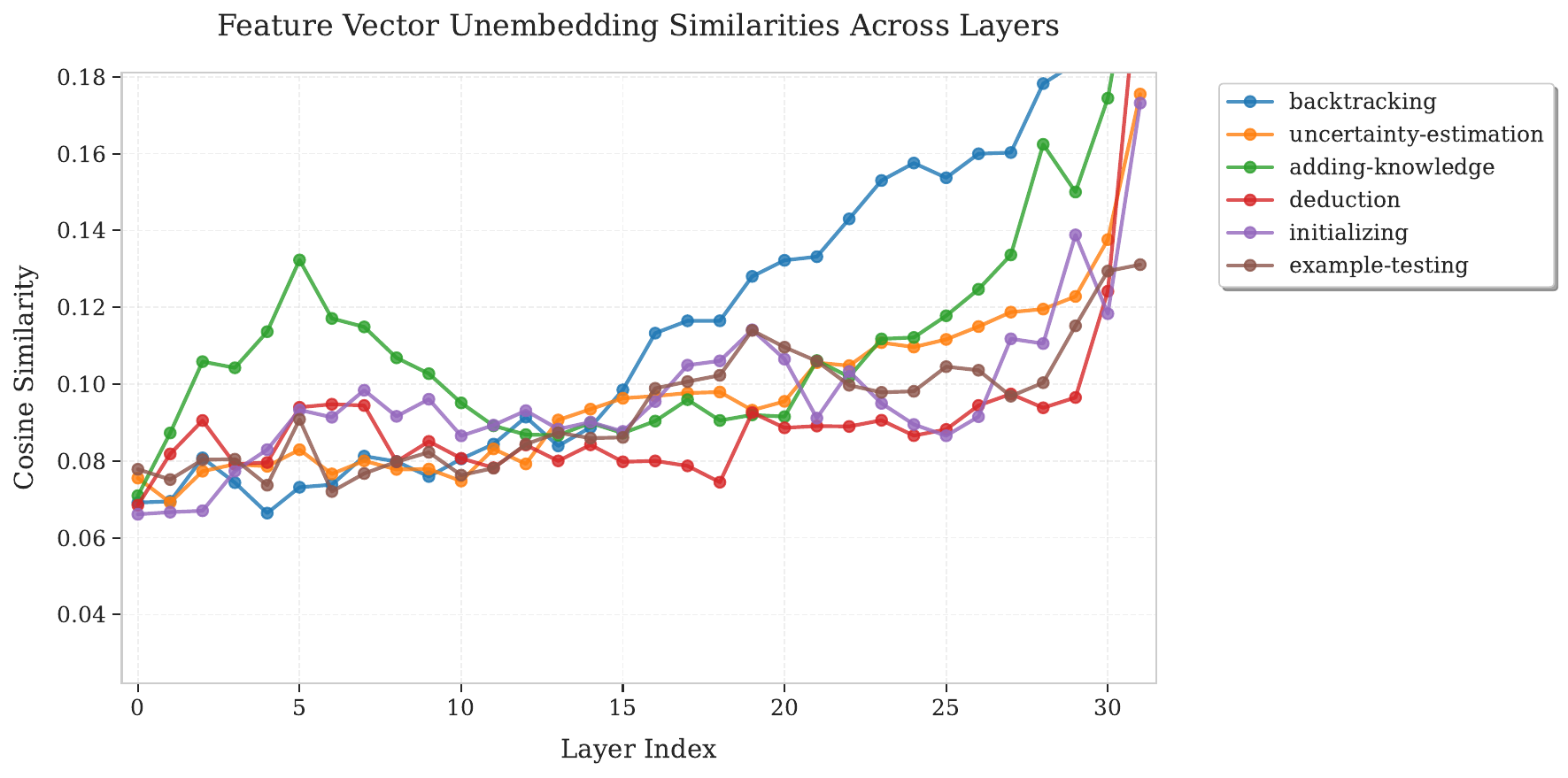}
        \caption{Unembedding similarities - Llama-8B}
        \label{fig:unembedding-similarity-llama}
    \end{subfigure}
    \hfill
    \begin{subfigure}[b]{0.49\linewidth}
        \centering
        \includegraphics[width=\linewidth]{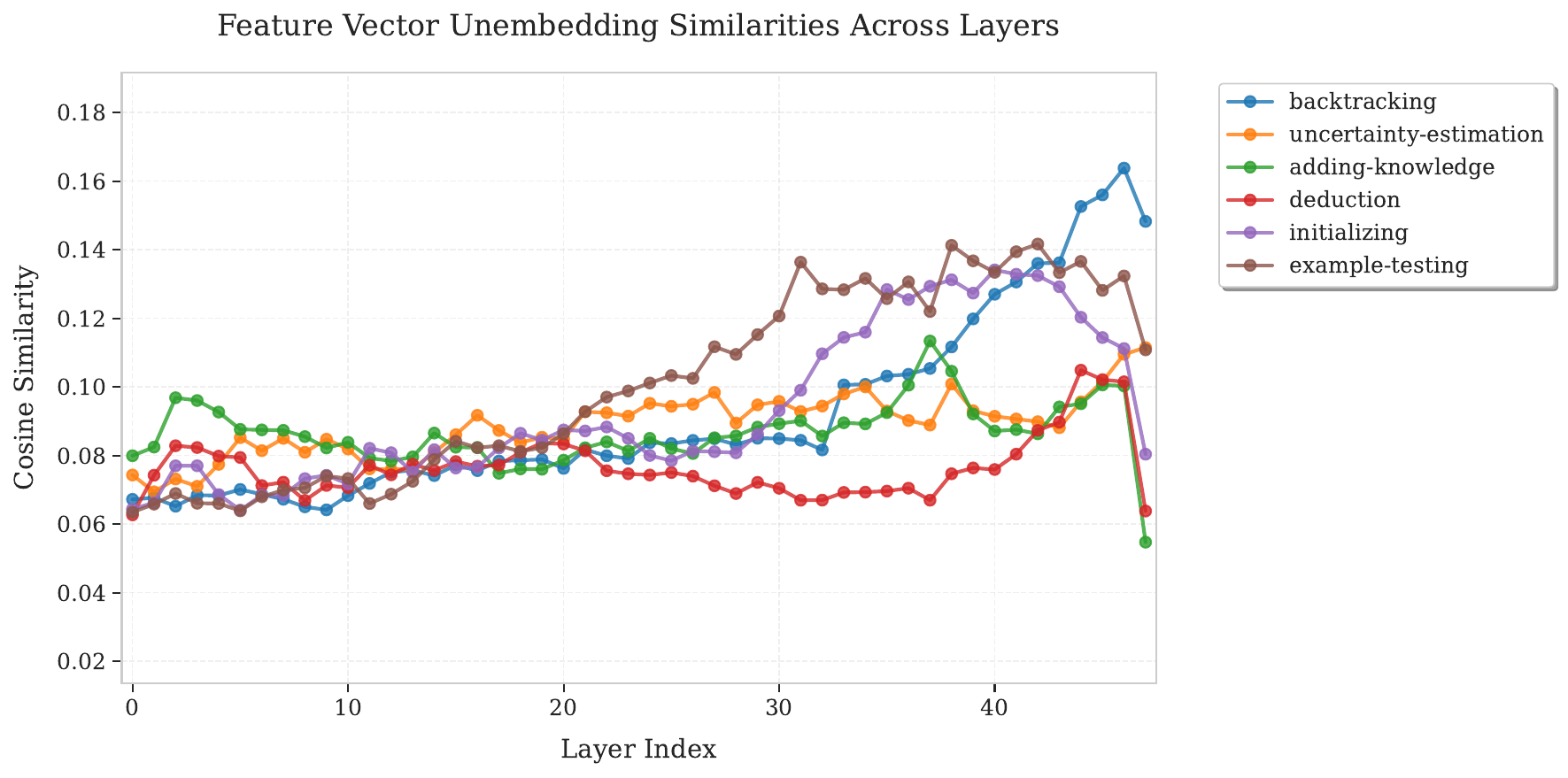}
        \caption{Unembedding similarities - Qwen-14B}
        \label{fig:unembedding-similarity-qwen}
    \end{subfigure}
    \caption{Layer-wise cosine similarities between steering vectors and embedding/unembedding matrices. The Llama model shows high embedding similarity in early layers, indicating strong correlation with token representations. This pattern is not observed in the Qwen model, explaining the different layer selection strategies for effective steering.}
    \label{fig:embedding-similarity}
\end{figure}

\cref{fig:embedding-similarity} shows the layer-wise similarity patterns for both models. The analysis reveals distinct architectural differences:

\textbf{DeepSeek-R1-Distill-Llama-8B}: Shows high similarity with embedding vectors in the first layers, explaining the high KL divergence observed at the beginning of the attribution plots (\cref{fig:attribution-effects}). This suggests that Llama retains token representation information in early layers, making steering vectors in these layers too correlated with specific tokens to be effective for behavioral control.

\textbf{DeepSeek-R1-Distill-Qwen-14B}: Does not exhibit this pattern, with more uniform similarities across layers. This architectural difference allows for more flexibility in layer selection for steering.

These findings inform our layer selection strategy: we prioritize layers with high KL divergence in attribution plots while avoiding early layers that show excessive correlation with embedding tokens. This approach ensures that our steering vectors capture behavioral patterns rather than token-specific representations.

\section{Steered Example (Adding Knowledge)}\label{app:example-steering}

\textbf{Task}:
\begin{lstlisting}[breakindent=0pt]
What's the probability of drawing exactly two hearts when dealing three cards from a standard deck?
\end{lstlisting}

\textbf{Original Response}:

\begin{lstlisting}[breakindent=0pt]
["initializing"]
Okay, so I need to figure out the probability of drawing exactly two hearts when dealing three cards from a standard deck.
["end-section"]

["uncertainty-estimation"]
Hmm, let's see.
["end-section"]

["adding-knowledge"]
I remember that probability problems often involve combinations,
["end-section"]

["deduction"]
so maybe I should start by recalling how combinations work.
["end-section"]

["adding-knowledge"]
First, a standard deck has 52 cards, right? And there are 13 hearts in the deck, one for each rank.
["end-section"]

["deduction"]
So, if I'm drawing three cards, I want exactly two of them to be hearts and one to be a non-heart. That makes sense.
["end-section"]

["deduction"]
I think the way to approach this is to calculate the number of favorable outcomes (i.e., drawing exactly two hearts and one non-heart) and then divide that by the total number of possible outcomes when drawing three cards.
["end-section"]

["adding-knowledge"]
So, probability is generally favorable over total.
["end-section"]

["initializing"]
Let me write that down: 
["end-section"]

["deduction"]
Probability = (Number of ways to choose 2 hearts and 1 non-heart) / (Total number of ways to choose 3 cards from the deck). Okay, so I need to compute both the numerator and the denominator.
["end-section"]

["deduction"]
Starting with the denominator, the total number of ways to choose 3 cards from 52.
["end-section"]

["adding-knowledge"]
That's a combination problem because the order doesn't matter when you're dealing cards.
["end-section"]

["adding-knowledge"]
The formula for combinations is C(n, k) = n! / (k! * (n - k)!), where n is the total number, and k is the number you're choosing.
["end-section"]

["deduction"]
So, the denominator is C(52, 3). Let me compute that: 
["end-section"]

["deduction"]
C(52, 3) = 52! / (3! * (52 - 3)!) = (52 * 51 * 50) / (3 * 2 * 1) = (52 * 51 * 50) / 6
["end-section"]

["initializing"]
Let me calculate that: 
["end-section"]

["uncertainty-estimation"]
52 divided by 6 is approximately 8.666..., but maybe I should compute it step by step.
["end-section"]

["deduction"]
First, 52 * 51 = 2652
["end-section"]

["deduction"]
Then, 2652 * 50 = 132600
["end-section"]

["deduction"]
Now, divide by 6: 132600 / 6 = 22100
["end-section"]

["deduction"]
So, the total number of ways is 22,100. Got that.
["end-section"]
\end{lstlisting}

\textbf{Positively Steered Response}

\begin{lstlisting}[breakindent=0pt]
["initializing"]
Okay, so I need to figure out the probability of drawing exactly two hearts when dealing three cards from a standard deck.
["end-section"] 

["uncertainty-estimation"]
Hmm, probability problems can be tricky,
["end-section"] 

["adding-knowledge"]
but I remember that probability often involves combinations and the concept of favorable outcomes over total possible outcomes.
["end-section"] 

["adding-knowledge"]
Let me start by recalling that a standard deck has 52 cards, divided into four suits: hearts, diamonds, clubs, and spades.
["end-section"] 

["adding-knowledge"]
Hearts are one of the four suits, and there are 13 cards in each suit.
["end-section"] 

["deduction"]
So, in a standard deck, there are 13 hearts, and the rest of the cards are non-hearts, which would be 52 minus 13, so 39 non-heart cards.
["end-section"] 

["adding-knowledge"]
When dealing with probabilities involving card draws, especially without replacement, the concept of combinations is really important.
["end-section"] 

["adding-knowledge"]
Combinations are used to determine the number of ways to choose a subset of items from a larger set, and the formula for combinations is C(n, k) = n! / (k! * (n - k)!), where n is the total number of items, and k is the number of items you're choosing.
["end-section"] 

["adding-knowledge"]
In this problem, we're dealing with hypergeometric distribution, which deals with successes and failures in draws without replacement.
["end-section"] 

["adding-knowledge"]
The hypergeometric probability formula is used to calculate the probability of k successes (in this case, drawing hearts) in n draws (here, n is 3 because we're dealing three cards).
["end-section"] 

["adding-knowledge"]
The formula is P(X = k) = (C(K, k) * C(N - K, n - k)) / C(N, n), where:
- N is the total number of items in the population, which in this case is the deck of 52 cards.
- K is the number of success states in the population, which here is the number of hearts, so K = 13.
- n is the number of draws, which is 3 for this problem.
- k is the number of observed successes, which is the number of hearts we want to draw. In this case, we're interested in exactly two hearts, so k = 2.
["end-section"] 

["deduction"]
So, for exactly two hearts in three card draws, we need to calculate the probability of drawing exactly 2 hearts and 1 non-heart card.
["end-section"] 

["deduction"]
The formula for this specific case would be: P(X = 2) = [C(13, 2) * C(39, 1)] /
["end-section"]
\end{lstlisting}

\textbf{Negatively Steered Response}

\begin{lstlisting}[breakindent=0pt]
["initializing"]
First, I need to determine the number of ways to draw exactly two hearts from the deck.
["end-section"]

["adding-knowledge"]
There are 13 hearts in the deck, and I want to choose 2 of them.
["end-section"]

["deduction"]
The number of ways to do this is 13 choose 2.
["end-section"]

["adding-knowledge"]
Then, I need to choose 1 card from the remaining 27 non-heart cards.
["end-section"]

["deduction"]
The number of ways to do this is 27 choose 1.
["end-section"]

["deduction"]
Finally, I multiply these two numbers together to get the total number of favorable outcomes.
["end-section"]

["deduction"]
Then, I divide by the total number of possible ways to draw 3 cards from the deck, which is 52 choose 3.
["end-section"]

["deduction"]
This gives me the probability of drawing exactly two hearts.
["end-section"]
\end{lstlisting}